\documentclass[sigconf, natbib]{acmart}

\usepackage{CJKutf8} 
\DeclareUnicodeCharacter{FF08}{(}
\DeclareUnicodeCharacter{FF09}{)}

\AtBeginDocument{%
  \providecommand\BibTeX{{%
    \normalfont B\kern-0.5em{\scshape i\kern-0.25em b}\kern-0.8em\TeX}}}

\acmConference[Preprint]{}{August}{2025}
\acmBooktitle{Preprint}

\renewcommand\footnotetextcopyrightpermission[1]{}
\usepackage{graphicx}
\usepackage{amsmath}
\usepackage{makecell}
\usepackage{array}
\usepackage{tablefootnote} 
\usepackage{subfigure}
\usepackage{adjustbox}
\usepackage{hyperref}

\begin{document}
\title{Multi-TW: Benchmarking Multimodal Models on Traditional Chinese Question Answering in Taiwan}


\author{Jui-Ming Yao}
\affiliation{%
  \institution{National Taiwan University of Science and Technology}
  \city{Taipei}
  \country{Taiwan}
}
\email{b11132009@mail.ntust.edu.tw}

\author{Bing-Cheng Xie}
\affiliation{%
  \institution{National Taiwan University of Science and Technology}
  \city{Taipei}
  \country{Taiwan}
}
\email{b11130009@mail.ntust.edu.tw}

\author{Sheng-Wei Peng}
\affiliation{%
  \institution{National Taiwan University of Science and Technology}
  \city{Taipei}
  \country{Taiwan}
}
\email{m11207330@mail.ntust.edu.tw}

\author{Hao-Yuan Chen}
\affiliation{%
  \institution{University of London}
  \city{London}
  \country{United Kingdom}
}
\email{hc118@student.london.ac.uk}

\author{He-Rong Zheng}
\affiliation{%
  \institution{National Taiwan University}
  \city{Taipei}
  \country{Taiwan}
}
\email{b10302350@ntu.edu.tw}

\author{Bing-Jia Tan}
\affiliation{%
  \institution{National Taiwan University}
  \city{Taipei}
  \country{Taiwan}
}
\email{b11115001@mail.ntust.edu.tw}

\author{Peter Shaojui Wang}
\affiliation{%
  \institution{National Taiwan University of Science and Technology}
  \city{Taipei}
  \country{Taiwan}
}
\email{shaojuiwang@mail.ntust.edu.tw}

\author{Shun-Feng Su}
\affiliation{%
  \institution{National Taiwan University of Science and Technology}
  \city{Taipei}
  \country{Taiwan}
}
\email{sfsu@mail.ntust.edu.tw}


\renewcommand{\shortauthors}{Yao, et al.}


\begin{abstract}
Multimodal Large Language Models (MLLMs) process visual, acoustic, and textual inputs, overcoming the limitations of single-modality LLMs. However, existing benchmarks often neglect tri-modal evaluation in Traditional Chinese and overlook inference latency. To fill this gap, we introduce \textbf{Multi-TW}, the first Traditional Chinese benchmark for evaluating the performance and latency of any-to-any multimodal models. Multi-TW comprises 900 multiple-choice questions (image \& text, audio \& text pairs) from authentic proficiency tests developed with the Steering Committee for the Test of Proficiency-Huayu (SC-TOP). We evaluated various any-to-any models and vision-language models (VLMs) with audio transcription. Our findings show closed-source models generally outperform open-source ones across modalities, though open-source models can excel in audio tasks. End-to-end any-to-any pipelines demonstrate significant latency advantages over VLM with separate audio transcription. Multi-TW offers a holistic view of model capabilities, underscoring the need for Traditional Chinese fine-tuning and efficient multimodal architectures.
\end{abstract}





\maketitle


\section{Introduction}
Pre-trained Large Language Models (LLMs), such as LLaMA~\citep{llama1, llama2} and Qwen~\citep{qwen1, qwen2, qwen2.5}, have demonstrated remarkable success across a wide range of natural language processing (NLP) tasks. However, these text-only models remain constrained by their single-modality input. To address this limitation, recent research has increasingly focused on Multimodal Large Language Models (MLLMs), which can jointly process and reason over visual, acoustic, and textual inputs~\citep{MLLM-survey1, MLLM-survey2}.

In the visual domain, models such as CLIP~\citep{clip} and Flamingo~\citep{flamingo} have shown that contrastive pretraining and multimodal fusion architectures enable state-of-the-art zero-shot image classification, image captioning, and few-shot visual reasoning~\citep{vlm-survey1, vlm-survey2}. Building upon these breakthroughs, Vision-Language Models (VLMs) like LLaVA~\citep{llava1.5} have pushed the frontier further, inspiring fine-tuned successors such as Vicuna~\citep{vicuna} and Alpaca~\citep{alpaca}, which expand multimodal reasoning capabilities across broader task domains. The models evaluated in our experiments, such as the LLaVA series, PaliGemma 2~\citep{paligemma2}, Idefics2~\citep{Idefics2}, Llama 3.2-Vision~\citep{Llama3.2-Vision}, UI-TARS~\citep{UI-TARS} and Qwen VL~\citep{qwen25vl} series, represent the cutting edge in these developments.


With the evolution of VLMs, increasing attention has turned toward audio-language modeling. Audio Language Models (ALMs) typically employ an audio encoder that transforms raw waveform signals into token representations that can be processed by a language model~\citep{ALM-survey1, ALM-survey2}. For instance, Qwen-Audio~\citep{qwen-audio} and Qwen-Audio2~\citep{qwen-audio2} utilize the Qwen model series~\citep{qwen1, qwen2} as their language modeling backbone and incorporate OpenAI’s Whisper~\citep{whisper} for end-to-end speech recognition. Other architectures, such as AudioPaLM~\citep{AudioPaLM}, fuse the text-based capabilities of PaLM-2~\citep{palm2} with the discrete audio token modeling of AudioLM~\citep{audiolm}, enabling both high-quality speech recognition and speech-to-speech translation in a unified framework.

More recently, research has progressed toward universal any-to-any multimodal models that support cross-modal input and output across vision, audio, and text. Prominent examples include NExT-GPT~\citep{nextgpt}, AnyGPT~\citep{anygpt} and Unified-IO 2 ~\citep{unifiedIO2}, all pushing the limits of unified multimodal intelligence. Later, this trend transfer into multilingual support, as shown in open-source models like Baichuan-Omni-1.5~\citep{Baichuan-Omni-1.5} and Qwen2.5-Omni~\citep{qwen25omni}, as well as closed-source systems such as gemini, which achieve strong performance in both Chinese and English understanding.


To rigorously evaluate the capabilities of such models, several benchmarks have been proposed. However, most evaluations still assess only two modalities at a time. For instance, NExT-GPT~\citep{nextgpt} and AnyGPT~\citep{anygpt} focus on pairwise modality evaluations. Recently, Qwen2.5-Omni~\citep{qwen25omni} and Baichuan-Omni-1.5~\citep{Baichuan-Omni-1.5} have adopted OmniBench~\citep{omnibench}, a tri-modal benchmark designed to assess performance across text, image, and audio simultaneously,
providing deeper insight into a model’s unified reasoning ability.


Despite these advances, a critical gap remains in the evaluation of multimodal models in Traditional Chinese. \textbf{Existing Traditional Chinese benchmarks are largely text-based.} TMMLU~\citep{tmmlu} and its extension TMMLU+~\citep{tmmlu+} provide comprehensive text-only evaluations of LLMs. VisTW~\citep{vistw} moves into the multimodal space by evaluating VLMs on multiple-choice and dialog-based tasks; however, no benchmark currently supports comprehensive evaluation across textual, visual, and acoustic modalities in Traditional Chinese. In addition to this linguistic gap, we observe that most existing benchmarks prioritize accuracy, often \textbf{overlooking model inference time}. This approach is insufficient for real-world applications where both accuracy and efficiency are crucial.

To address this gap, we introduce \textbf{Multi-TW}, the first benchmark specifically designed for evaluating the performance and latency of any-to-any multimodal models in Traditional Chinese. Multi-TW consists of image-text and audio-text pairs, enabling rigorous evaluations that cover textual, visual, and acoustic modalities. All datasets are open-sourced and available at: \textbf{\url{https://drive.google.com/drive/folders/1IvBOXR1GpMNtst0T3HT6dM59_ASlXdyn}}.

In summary, our contributions are as follows:
\begin{itemize}
    \item We propose \textbf{Multi-TW}, the first Traditional Chinese benchmark for rigorous evaluation across text, audio, and visual inputs.
    \item We collaborated with the Steering Committee for the Test of Proficiency-Huayu to incorporate authentic, real-world assessment tasks into our machine evaluation framework.
    \item We conduct comprehensive experiments on both any-to-any models and VLMs (the latter using ASR for audio input).
    \item In addition to accuracy, we evaluate latency to offer a more holistic view of model performance in real-world settings.
\end{itemize}

\section{Multi-TW Benchmark}
In this section, we provide a concise overview of Multi-TW, including its construction process, validation procedures, and data format specifications to support reproducibility. Our dataset is derived from real-world exams, detailed further in Section \ref{sec:data_analysis}.

\subsection{Data Construction}
To construct the Multi-TW dataset, we collaborated with the Steering Committee for the Test of Proficiency-Huayu (SC-TOP), a dedicated agency responsible for developing and promoting Taiwan’s Mandarin proficiency tests for non-native speakers. These exams, primarily in a multiple-choice format, underwent rigorous utility analysis to ensure their practical value and effectiveness.

    \begin{figure}[ht]
        \centering
        \includegraphics[width=0.95\linewidth]{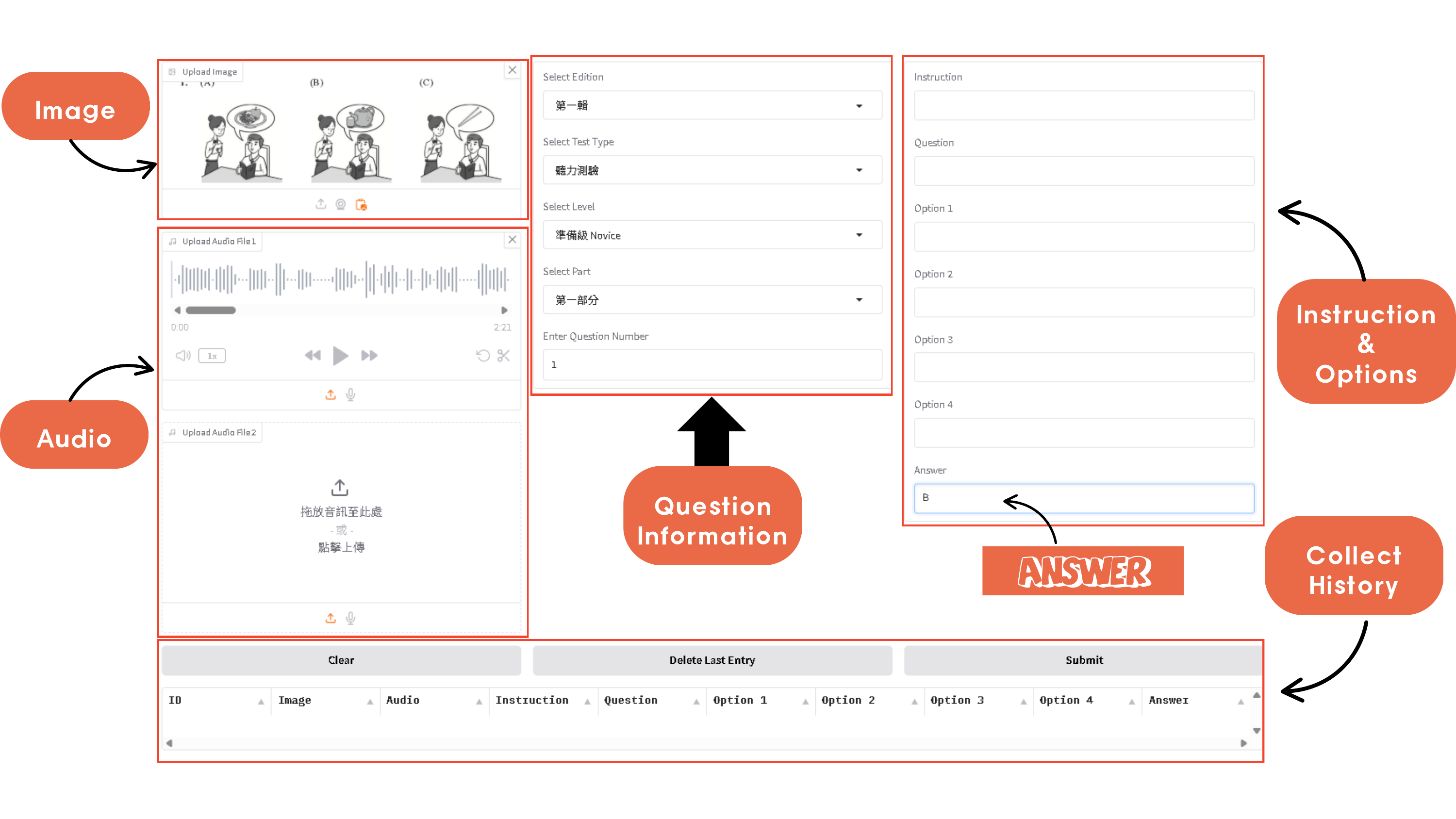} 
        \caption{Illustration of data collection interface.}
        \label{figure:collection_ui}
    \end{figure}

\textbf{Construction}.
The construction phase spanned from September 2023 to December 2023, primarily using publicly available data. All items in Multi-TW underwent a standardized collection and processing workflow performed by our research team to ensure consistency and accuracy. We developed an interface to accelerate data collection and automate labeling, as depicted in Figure~\ref{figure:collection_ui}. Initially, purely textual questions were removed. The remaining items, which involved various combinations of modalities, were then curated to form image-text and audio-text pairs. To address data imbalance and expand the image-text subset, some questions originally coupling image and audio were adapted by extracting their ground-truth audio transcripts, which were then paired with the corresponding image as the textual component. Subsequently, each image-text and audio-text multiple-choice item was serialized into a unified JSON schema, containing the original question, response options, instructions, and references to the separately stored image or audio files. During construction, instructions were classified into seven categories to prevent excessive fragmentation of task types. Further details are depicted in Table~\ref{table:json_fields}.

\textbf{Quality Control}.
To ensure data integrity, each image-text and audio-text pair was independently reviewed by a second annotator to verify content consistency and accuracy, ensuring the absence of syntax errors, missing information, or incorrect answer choices. Our quality control process involved multiple stages:
    \begin{enumerate}
        \item \emph{Completeness Check:} Annotators inspected each question to confirm the presence of all required components: text (prompt, options, and solution index), image or audio file, and associated metadata. Entries with missing or inconsistent elements (e.g., a mismatched file name) were flagged and corrected.
        \item \emph{File Consistency Check:} Each image was viewed to confirm it was properly formatted (150 dpi PNG), and each audio clip was played to ensure audible clarity in the specified 128 kbps MP3 (or other, specify format) setting. Invalid or corrupted files were replaced or re-processed.
        \item \emph{Label Accuracy Verification:} Given that the dataset tests language proficiency, annotators carefully matched the text content with the corresponding image or audio. For the image-text subset, the visual context had to align with the question stem and options (e.g., an illustration of a given scenario). For audio-text items, the spoken content was compared with the multiple-choice options to verify that the designated answer was correct.
        \item \emph{Final Confirmation:} After all corrections were made, each question was subjected to a final review to verify that the files and metadata were correctly updated. Only after passing this final check was the question approved for inclusion in the final dataset.
    \end{enumerate}

\textbf{Reproducibility}. To ensure reproducibility and facilitate data management, a system was developed to encode the original source location of each data item. The following details the significance of the dataset identifier. Files are organized into directories named by a five-part identifier, read from left to right as follows:
    \begin{enumerate}
        \item Volume (1–5): The 'booklet' number of the data source.
        \item Section (L/R): L for Listening, R for Reading.
        \item Level (N/A/B/C): A difficulty rating where N denotes “Novice” and A, B, and C correspond to Bands A, B, and C, respectively.
        \item Part (Pn): The part number within the section (e.g., P1, P2).
        \item Question: The index of the question within that part.
    \end{enumerate}
This structure facilitates verification and aids researchers in data retrieval and collection. More detailed information on the data format is provided in Table \ref{table:json_fields}.

    \begin{table}[htbp]
        \centering
        \caption{Key Fields in the JSON Annotation Format.}
        \label{table:json_fields}
        \begin{tabular}{l p{0.7\columnwidth}}
            \toprule
            Field Name & Description \\
            \midrule
            id           & Unique item identifier (e.g., "01-L-A-P1-001"). \\
            image        & Relative image file path (string or null). \\
            audio        & Relative audio file path (string or null). \\
            instruction  & Instructions for the question. \\
            question     & Textual content of the question. \\
            options      & List of options, typically prefixed with (A)–(D) (e.g., ["(A) Option 1", "(B) Option 2"]). \\
            answer       & Correct option identifier: "A", "B", "C", or "D". \\
            \bottomrule
        \end{tabular}
    \end{table}

\subsection{Data Analysis}
\label{sec:data_analysis}

    \begin{figure*}[ht]
        \centering
        \includegraphics[width=0.95\linewidth]{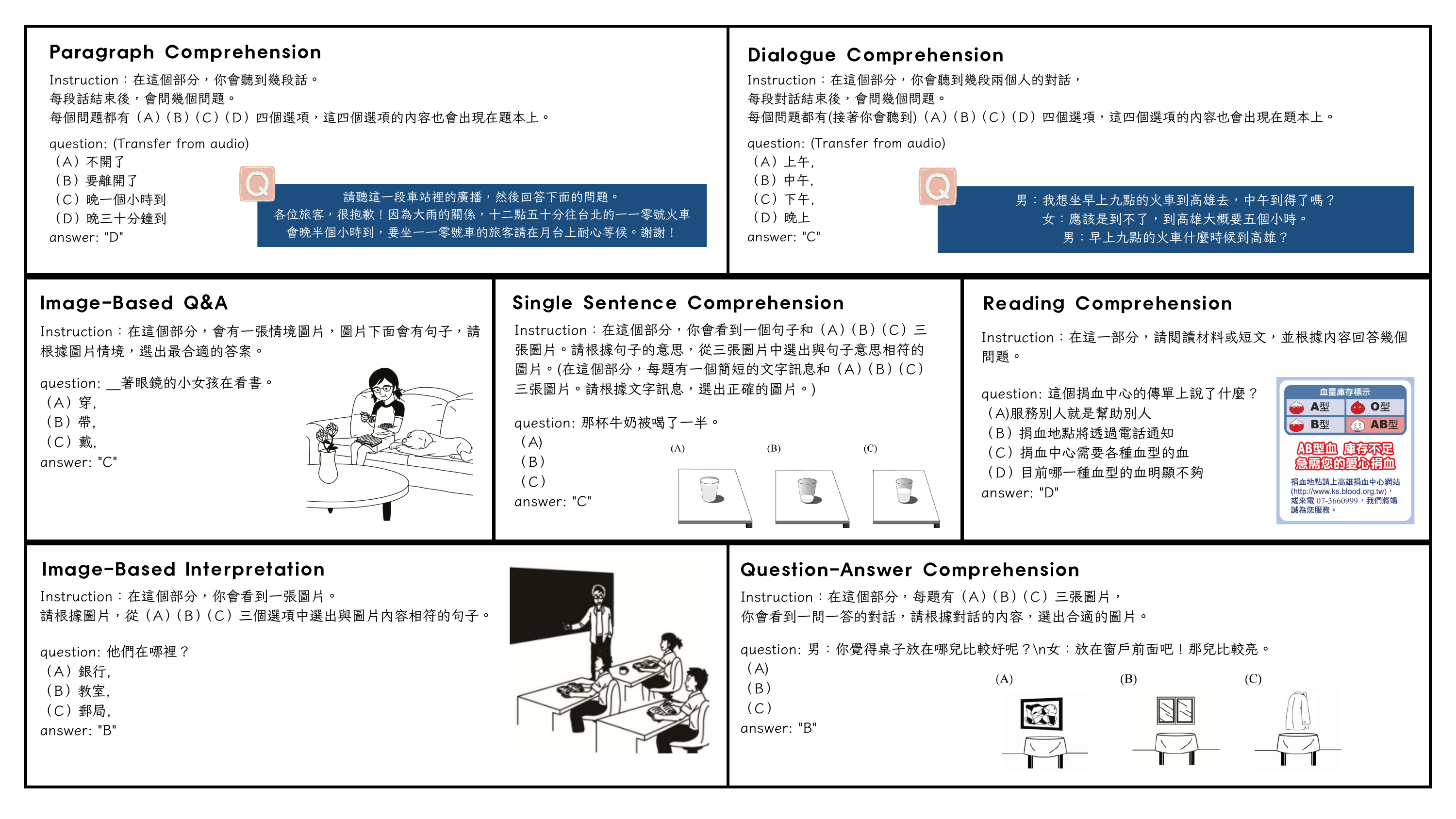} 
        \caption{Illustration of samples from the Multi-TW dataset.}
        \label{figure:Overview_sample}
    \end{figure*}

This section provides a detailed analysis of Multi-TW to facilitate a comprehensive understanding of its characteristics, including its size and the distribution of question types and modalities. Through this analysis, our objective is to concretely characterize our dataset and highlight its distinctions from existing benchmarks. Finally, we compare Multi-TW against other established datasets to underscore its unique characteristics and strengths.

\textbf{Data Size.}
Multi-TW comprises 900 multiple-choice questions curated to assess Traditional Chinese proficiency in a multimodal context. The dataset is equally divided into 450 image-text items and 450 audio-text items. In the following sections, we refer to these as 'vision-based items' and 'audio-based items,' respectively. This balanced design enables direct comparison of model performance on visual versus auditory modalities paired with Traditional Chinese text and encourages the development of models that handle both input types proficiently.

\begin{figure}[ht]
        \centering
        \includegraphics[width=0.97\linewidth]{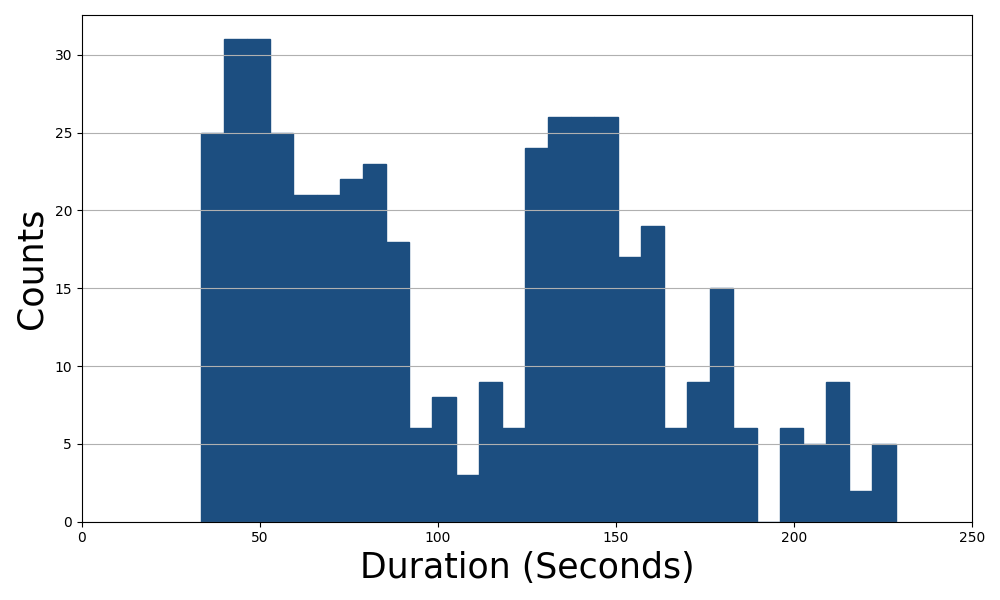} 
        \caption{Distribution of audio durations in Multi-TW.}
        \label{figure:time_distribution}
    \end{figure}

\textbf{Data Distribution.}
The vision-based subset features 397 distinct images and includes 407 three-choice items alongside 43 four-choice items. These images depict contextual illustrations, diagrams, and real-world scenarios. All audio-based items employ a four-choice format. Consequently, the 900-item benchmark comprises 407 three-choice questions and 493 four-choice questions (43 from vision-based and 450 from audio-based). For the audio-based items, the average question length is approximately 12 words, and the average option length is approximately 10 words. The average duration of the audio is approximately 107.5 seconds, as illustrated in Figure~\ref{figure:time_distribution}.

\begin{figure}[ht]
        \centering
        \includegraphics[width=0.95\linewidth]{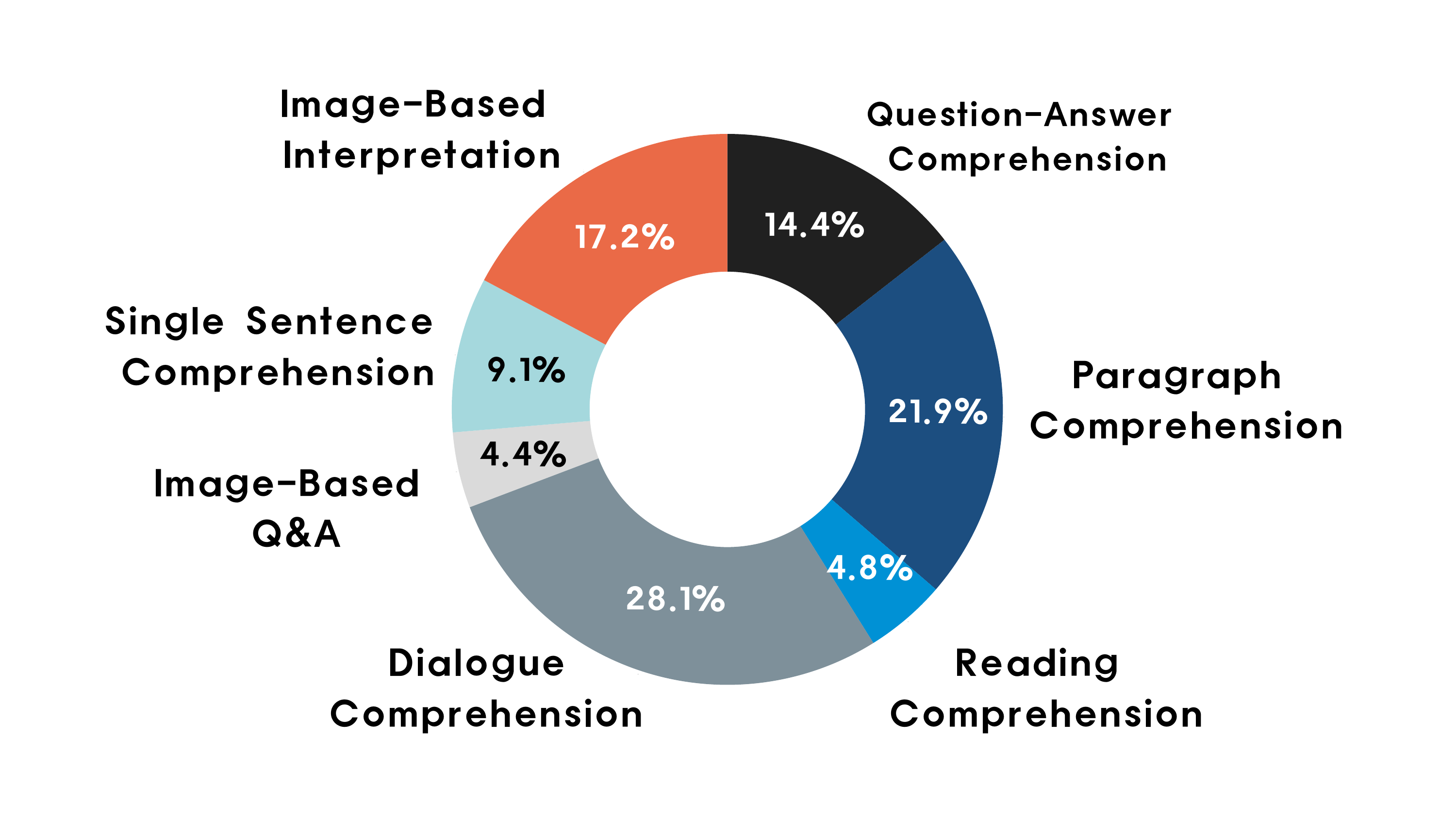} 
        \caption{Distribution of question types in Multi-TW.}
        \label{figure:Catogory_chart}
    \end{figure}

\textbf{Task Formulation.}
Multi-TW evaluates multimodal understanding by measuring performance on two primary task types: vision-based tasks and audio-based tasks. These are structured as multiple-choice questions (MCQs), namely Vision-based MCQ and Audio-based MCQ.
\begin{itemize}
  \item \textbf{Audio-based MCQ} comprises two subtasks:
    \begin{itemize}
      \item Dialogue Comprehension
      \item Passage Comprehension
    \end{itemize}
  \item \textbf{Vision-based MCQ} comprises five subtasks:
    \begin{itemize}
      \item Dialogue Comprehension
      \item Image Comprehension
      \item Reading Comprehension
      \item Sentence-to-Image Matching
      \item Image-to-Sentence Matching
    \end{itemize}
\end{itemize}
Details of subtask distribution are provided in Figure~\ref{figure:Catogory_chart}. This diverse mix of task types ensures that Multi-TW evaluates a broad spectrum of multimodal understanding capabilities.

\textbf{Comparison with Existing Benchmarks.}
Table \ref{table:compare_other} presents a comparison of Multi-TW with other notable Traditional Chinese language evaluation datasets. While existing benchmarks like TMMLU+~\citep{tmmlu+} focus on text-only LLM capabilities, and VisTW-MCQ~\citep{vistw} and ALM-Bench~\citep{ALM-Bench} incorporate vision and text, Multi-TW, to the best of our knowledge, is the first benchmark to provide comprehensive image-text and audio-text evaluation for Traditional Chinese, thereby covering visual, textual, and auditory modalities. By unifying these input types within a single benchmark framework for Traditional Chinese, it fills a critical gap and enables a more holistic evaluation of multimodal models. Moreover, beyond its rich modality and linguistic features, Multi-TW's audio samples average 107.5 seconds in length, substantially longer than the 9.12 seconds typical of OmniBench~\citep{omnibench} (which primarily tests English). This extended duration enables a more rigorous evaluation of long-form listening comprehension abilities.

    \begin{table}[htbp]
      \centering
      \caption{Comparison of Multi-TW with other datasets. For ALM-Bench, we only compare the subset for Traditional Chinese. (A: audio, T: text, V: vision) (Traditional Chinese: zh, English: en)}
      \label{table:compare_other}
      \begin{tabular}{lrcrr}
        \toprule
        Dataset & Modalities & Language & Test size & Subjects \\
        \midrule
        TMMLU+~\citep{tmmlu+} & T & zh & 20,118 & 66 \\
        ALM-Bench~\citep{ALM-Bench} & T, V & zh & 52 & 13 \\
        VisTW-MCQ~\citep{vistw} & T, V & zh & 3,795 & 21 \\
        OmniBench~\citep{omnibench} & A, T, V & en & 1,142 & 8 \\
        \textbf{Multi-TW (Ours)} & A, T, V & zh & 900 & 7 \\
        \bottomrule
      \end{tabular}
    \end{table}

\section{Experiments}
To demonstrate the utility of Multi-TW and establish initial performance benchmarks, we conducted experiments using a variety of publicly available multimodal language models. This section details our experimental setup, the models evaluated, and the observed results.

\subsection{Experiment Setup}
All experiments were conducted on an NVIDIA A100-SXM4 80GB GPU. All 900 questions in Multi-TW were used for evaluation in a zero-shot setting. The evaluation metric reported is exact-match accuracy, reflecting the percentage of correctly answered multiple-choice questions. We detail our prompting strategy, answer extraction, and time measurement protocols below.

\textbf{Prompting Strategy.}
For all evaluated models, a uniform prompt was appended to each question. The general prompt template provided to the models is as follows (in Traditional Chinese):
\begin{CJK}{UTF8}{bsmi}
    \begin{quote}
        \texttt{\{question\}}\\
        \texttt{僅輸出正確答案的字母，格式必須為 'A', 'B', 'C', 'D'，輸出限制為單個字母，無需解釋。}
    \end{quote}
\end{CJK}

This prompt instructs the model to directly output a single character representing the chosen option, without any additional explanation or reasoning.

\textbf{Answer Extraction.}
To extract answers from the model-generated content, we constrained the model’s output to a single token and applied a regular expression to capture the first occurrence of 'A', 'B', 'C', or 'D'. If the regular expression failed to retrieve a valid response, a fallback mechanism was implemented where a random option from the available choices was selected. This ensures consistent answer provision across all evaluations.

\textbf{Time Measurement.} We recorded the elapsed time for four sequential stages: data loading, data preprocessing, model inference, and metric computation. Data preprocessing and model inference account for the majority of runtime and utilize identical code across all open-source models. Therefore, our timing analysis focuses primarily on the combined duration of these two phases for open-source models. Closed-source models were omitted from this specific latency analysis, as their response times are dominated by external API calls and network transmission, which are not directly comparable. To eliminate variability from differing output lengths, we fixed the model’s maximum generation length to one token for all timed experiments.

\subsection{Model Selection}
We evaluated several any-to-any models that process text, image, and audio inputs to generate text output in Traditional Chinese, as well as several VLMs where audio input was provided via ASR transcripts. These models, presented in Tables \ref{table:anytoany_result} and \ref{table:vlms_result}, span both closed- and open-weight categories and were selected based on their state-of-the-art performance, availability, architectural diversity, and varying degrees of exposure to Chinese language data.
For closed-source any-to-any models, we selected gemini-2.0-flash and gemini-1.5-flash from Google. For open-source any-to-any models, we chose the Qwen2.5-Omni series and Baichuan-Omni-1.5, both pretrained primarily on Simplified Chinese. Although Simplified and Traditional Chinese share lexical similarities, they differ substantially in character forms and orthographic conventions. We also incorporated UnifiedIO-2, an encoder-decoder Transformer pretrained from scratch mostly on English data (with a small multilingual fraction from mC4~\citep{mc4}), making it a useful test for zero-shot cross-script transfer as it has not been specifically fine-tuned for either Chinese variant.
For VLMs, we employed Whisper-large~\citep{whisper} to transcribe audio inputs into text for the audio-text tasks. The selected VLMs include Qwen2.5-VL-7B, Qwen2-VL-7B, Llama-3.2-11B-Vision, UI-TARS-1.5-7B, Idefics2-8b, the LLaVA series, and PaliGemma2. This selection reflects the current landscape and provides a broad overview of VLM capabilities on our benchmark.

    \begin{table*}[htbp]
      \centering
      \caption{Performance of Any-to-Any Multimodal Models on Multi-TW.}
      \label{table:anytoany_result}
      \begin{tabular}{lcccc}
        \toprule
        \textbf{Models} & \textbf{Overall Acc.} & \textbf{Image-Text Acc.} & \textbf{Audio-Text Acc.} & \textbf{Inference Time (s)} \\
        \midrule
        gemini-2.0-flash  & \textbf{0.8900} & \textbf{0.8800} & \textbf{0.9000} & - \\
        gemini-1.5-flash & \underline{0.8111} & \underline{0.7644} & 0.8578 & - \\
        Qwen2.5-Omni-7B & 0.6534 & 0.4156 & \underline{0.8911} & 744 \\
        Baichuan-Omni-1.5 & 0.6289 & 0.4822 & 0.7756 & \underline{569} \\
        Qwen2.5-Omni-3B & 0.5878 & 0.3377 & 0.8378 & 712 \\
        UnifiedIO-2-XL & 0.2589 & 0.2600 & 0.2578 & \textbf{467} \\
        \bottomrule
      \end{tabular}
    \end{table*}

    \begin{table*}[t]
      \centering
      \caption{Performance of Vision-Language Models (VLMs) with ASR (Whisper-large) on Multi-TW.}
      \label{table:vlms_result}
      \begin{tabular}{lcccc}
        \toprule
        \textbf{Models} & \textbf{Overall Acc.} & \textbf{Image-Text Acc.} & \textbf{Audio Transcript-Text Acc.} & \textbf{Inference Time (s)} \\
        \midrule
        Qwen2.5-VL-7B-Instruct & \textbf{0.8423} & \textbf{0.8267} & \textbf{0.8578} & 1216 \\
        Qwen2-VL-7B-Instruct   &  \underline{0.8033} &  \underline{0.7822} & 0.8244 & \textbf{1187} \\
        UI-TARS-1.5-7B & 0.7823 & 0.7378 &  \underline{0.8267} & 2131 \\
        Llama-3.2-11B-Vision-Instruct & 0.5578 & 0.4711 & 0.6444 & 1308 \\
        idefics2-8b   & 0.4167 & 0.5156 & 0.3178 & 1228 \\
        llava-v1.6-mistral-7b & 0.4100 & 0.4178 & 0.4022 & 1305 \\
        llava-v1.6-vicuna-7b  & 0.3345 & 0.4022 & 0.2667 & 1302 \\
        llava-v1.5-7b          & 0.3211 & 0.3911 & 0.2511 &  \underline{1201} \\
        paligemma2-10b-pt-896      & 0.2600 & 0.2800 & 0.2400 & 1727 \\
        \bottomrule
      \end{tabular}
    \end{table*}

\section{Results and Analysis}
This section offers a summary of performance across all evaluated models on the 900-item Multi-TW benchmark, comparing accuracy on the image-text and audio-text subsets alongside inference latency.

\textbf{Performance on Any-to-Any Models.}
Table~\ref{table:anytoany_result} illustrates the results for any-to-any models across overall accuracy, image-text subset accuracy, audio-text subset accuracy, and inference time. Key observations include:
1) The Qwen2.5-Omni series and Baichuan-Omni-1.5, despite being primarily pretrained and fine-tuned on Simplified Chinese, achieve competitive accuracy on Traditional Chinese inputs, particularly on audio-text tasks.
2) In contrast, UnifiedIO-2-XL, with limited exposure to Chinese, often failed to produce meaningful answers. Manual inspection of its responses (when constraining output length to 30 tokens) revealed that in 78 cases the model echoed the first option’s Chinese description, and in 807 cases it consistently selected option “A.”
3) Qwen2.5-Omni-7B exhibited the longest inference time among the open-source any-to-any models, approximately 30.8\% longer than Baichuan-Omni-1.5 (11B parameters). This suggests that parameter count is not the sole determinant of inference speed.
4) The results reveal a significant performance gap between open-source and closed-source models, especially in the image-text domain, highlighting the urgent need for dedicated Traditional Chinese fine-tuning and more robust vision components in open-source any-to-any models.

\textbf{Performance on Vision Language Models (with ASR).}
We evaluated a range of VLMs using Whisper-large for audio transcription. Table~\ref{table:vlms_result} reports overall accuracy, image-text accuracy, audio-transcript-text accuracy, and inference time. Key observations are:
1) Qwen2.5-VL-7B-Instruct and UI-TARS-1.5-7B lead among the evaluated VLMs. The competitive results from these models, developed by organizations with a strong focus on Chinese AI, suggest that extensive pre-training on relevant Chinese-language corpora is a crucial factor for strong performance.
2) In contrast, models like Llama-3.2-11B-Vision-Instruct, despite their large parameter counts or general multimodal capabilities, exhibit notably lower performance, potentially due to less exposure to Traditional Chinese data or specific task alignments.

\textbf{Performance on Latency.}
Open-source any-to-any models completed inference in a range of 467–744 seconds for the entire 900-item benchmark. In comparison, VLMs coupled with an ASR pipeline (Whisper-large for audio transcription, then VLM for comprehension) required 1,187–2,131 seconds, reflecting the overhead of the two-stage processing for audio-related tasks. In addition, while closed-source models' runtimes are not directly comparable due to API encapsulation, they generally exhibit higher end-to-end latency in practice for batch processing due to network factors, though individual query latency might be low.

\section{Conclusion and Future Work}
To address the gap in evaluating Multimodal Large Language Models capable of processing visual, acoustic, and textual inputs, particularly in Traditional Chinese, we introduced \textbf{Multi-TW}, the first benchmark of its kind. This dataset provides new insights into current multimodal large language models' abilities, including their performance and latency on Traditional Chinese tasks.
Our evaluation reveals that while closed-source models generally achieve strong performance across both image and audio modalities, open-source alternatives currently tend to perform better on audio-text tasks compared to image-text tasks when using any-to-any architectures. The VLM plus ASR approach can achieve strong results but incurs higher latency for audio tasks. We also found that end-to-end any-to-any models offer notable latency advantages over cascaded VLM plus ASR pipelines for processing audio inputs. Our findings underscore the need for more appropriate architecture designs and targeted fine-tuning data for robust multimodal integration, especially for Traditional Chinese.

In future work, we will examine how cross-lingual transfer capabilities influence the performance of Simplified Chinese-trained models on Traditional Chinese reasoning tasks. We also plan to evaluate latency under more rigorous, parallelized experimental conditions and explore alternative settings, such as streaming inference. Furthermore, expanding Multi-TW to include generative tasks and more complex reasoning scenarios will be a key direction.


\begin{acks}
    The authors would like to thank Steering Committee for the Test Of Proficiency-Huayu for agreeing to release the Test of Chinese as a Foreign Language data for this study. The responsibility for errors in fact or judgment is ours. We also extend our gratitude to Professor Yun-Nung Chen from National Taiwan University for her invaluable guidance and support.
\end{acks}



\bibliographystyle{ACM-Reference-Format}
\bibliography{sample-base}

\end{document}